\documentclass{article}
\usepackage[a4paper, total={6in, 9in}]{geometry}
\usepackage[utf8]{inputenc}
\usepackage{hyperref}
\usepackage{graphicx}
\usepackage{amsmath}
\usepackage{subfig}
\usepackage{tikz}
\usepackage{color}
\usepackage{wrapfig}
\usetikzlibrary{matrix}

\definecolor{darkgreen}{rgb}{0,0.5,0}

\title{Concepts, Properties and an Approach for\\Compositional Generalization}
\author{Yuanpeng Li}
\date{}

\begin{document}

\maketitle

\begin{abstract}
Compositional generalization is the capacity to recognize and imagine a large amount of novel combinations from known components.
It is a key in human intelligence, but current neural networks generally lack such ability.
This report connects a series of our work for compositional generalization, and summarizes an approach.
The first part contains concepts and properties.
The second part looks into a machine learning approach.
The approach uses architecture design and regularization to regulate information of representations.
%
This report focuses on basic ideas with intuitive and illustrative explanations.
We hope this work would be helpful to clarify fundamentals of compositional generalization and lead to advance artificial intelligence.
\end{abstract}

\section{Introduction}
\begin{figure}[!t]
\centering
\subfloat[
Sphinx
]{
\includegraphics[height=0.3\textwidth]{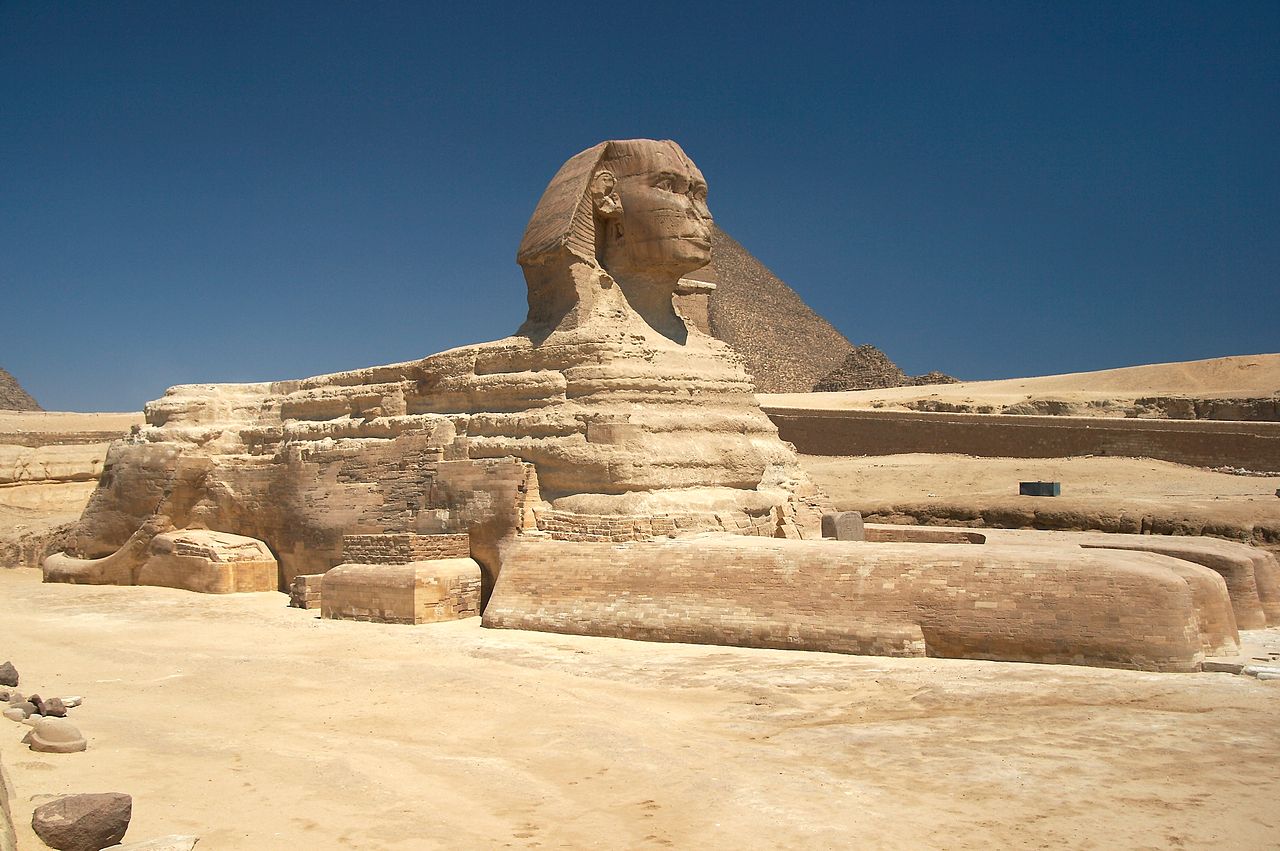}
\label{fig:sphinx}
}
\quad\quad
\subfloat[
Centaur
]{
\includegraphics[height=0.3\textwidth]{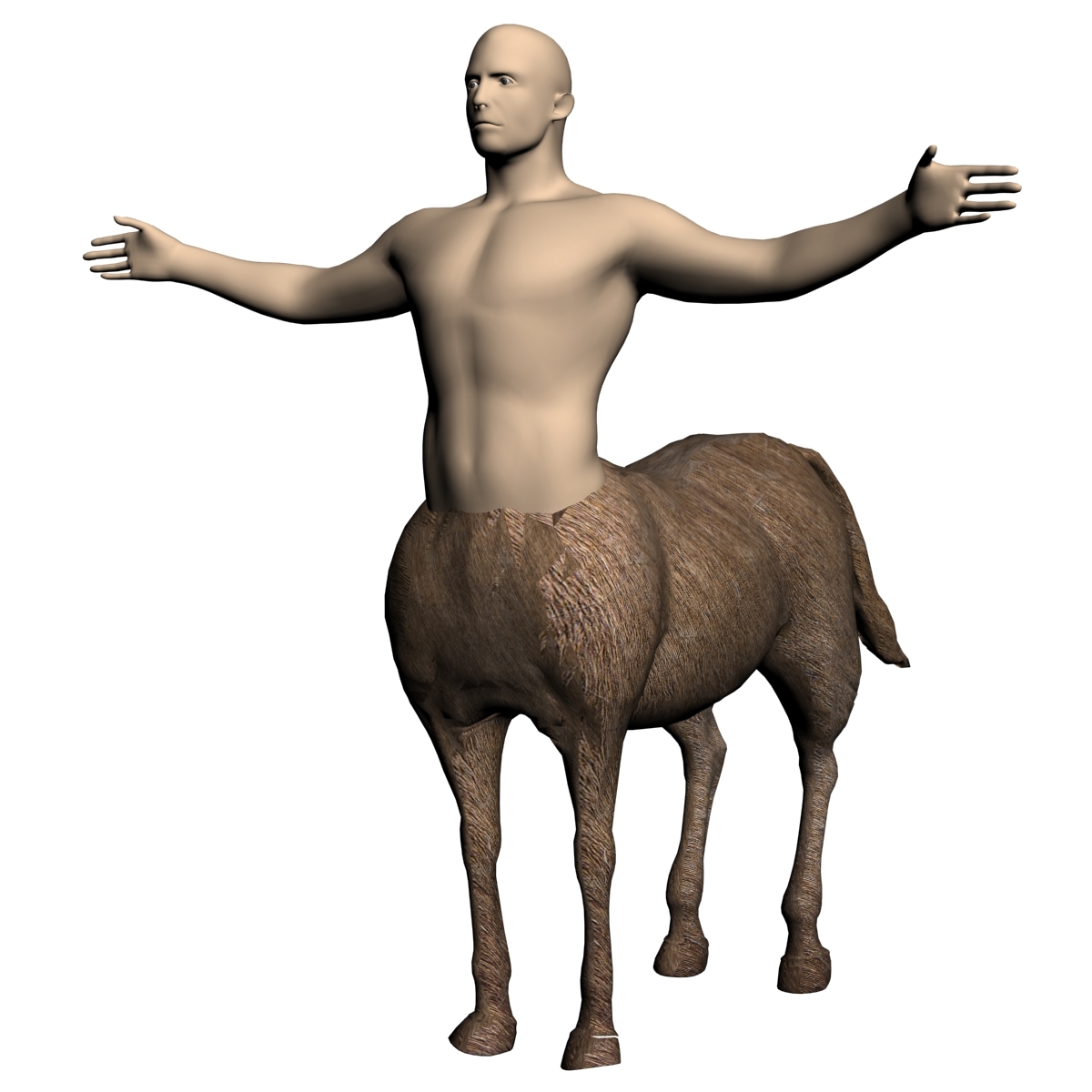}
}
\caption{
    Recombine parts of bodies to create an imagination.
}
\label{fig:example_myth}
\end{figure}

Humans leverage compositional generalization to recombine familiar concepts to understand and create new things.
We have been using this ability from early civilization.
For example, Sphinx has the face of a human and the body of a lion (Figure~\ref{fig:sphinx}).
We do not see such a living animal, but ancient people could create it and we can recognize it.
This shows we are able to recombine different parts of seen objects for an unseen object.
%
Sphinx actually also has wings of an eagle, and the type of wings can be another part to combine.
This means we have an exponentially large amount of combinations as the number of parts grows.
So compositional generalization helps humans to efficiently learn from a few training data, and generalize to a many unseen combinations.
We hope machines also have such ability.




Different compositional generalization approaches have been investigated, such as architecture design~\cite{russin2019compositional,goyal2019recurrent},
independence assumption~\cite{higgins2017beta,burgess2018understanding},
data augmentation~\cite{andreas-2020-good,akyurek2020learning},
causality~\cite{bengio2020a,li2021efficiently},
reinforcement learning~\cite{liu2020compositional},
group theory~\cite{Gordon2020Permutation} and
meta-learning~\cite{lake2019compositional}.
There are also general discussions~\cite{bengio2017consciousness,goyal2020inductive}.
Compositional generalization has been applied in many areas, such as instruction learning~\cite{lake2019compositional}, grounding~\cite{ruis2020benchmark}, continual learning~\cite{jin2020visually},
question answering~\cite{andreas2016,hudson2019gqa,keysers2020measuring}, reasoning~\cite{talmor2020teaching}, zero-shot learning~\cite{sylvain2020locality} and language inference~\cite{geiger-etal-2019-posing}.
In this report, we focus on summarizing a series of our work for theoretical discussions~\cite{li2021gradient,li2021transferability,li2021necessary}.
Please refer to these papers for concrete examples~\cite{li2019compositional,li2021grounded} and applications~\cite{Li2020Compositional}.
Please also find broad related work in the papers.

In this report, we first discuss concepts and properties related to compositional generalization.
Based on them, we clarify the setting in our scope (Section~\ref{sec:settings}).
We then propose an approach with architecture design, training and inference.
We also share conjectures to partially explain some human behaviors, such as system 1 and system 2 cognition.
This report has three main key points.
First, what is \textbf{compositional generalization}.
Second, what is \textbf{conditional independence property}, and how does it help compositional generalization.
Third, how to \textbf{control random variable information}, and how does it enable conditional independence property.
We will explain them in the following sections and summarize them in conclusion.

\section{Concepts and properties}
In this section, we first introduce compositional generalization and disentangled representation.
We then discuss two questions about disentangled representation: subjectivity of components and conditional independence property.


\subsection{Compositional generalization}
\begin{figure}[!ht]
\centering
\includegraphics[width=0.8\textwidth]{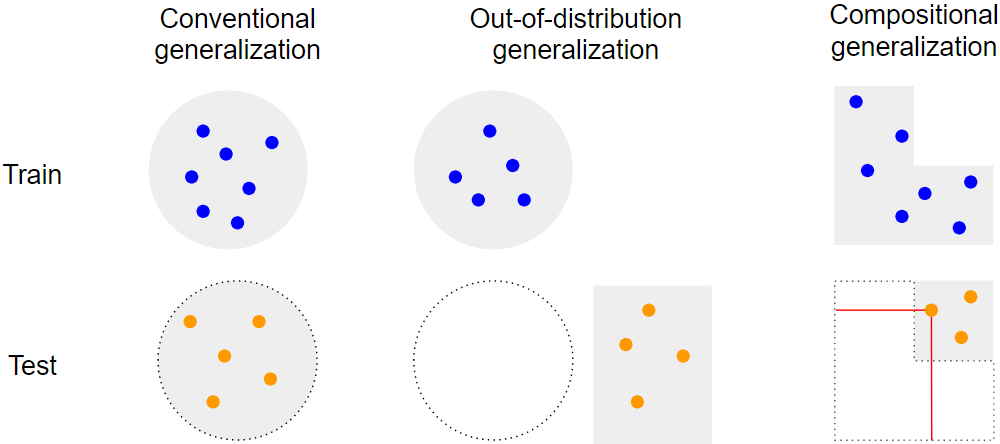}
\caption{
Types of generalization. Gray areas are input distributions (manifolds).
}
\label{fig:generalization_types}
\end{figure}

We compare different types of generalization to describe compositional generalization.
The images in Figure~\ref{fig:generalization_types} only contain input distributions for explanation purposes.
Conventionally, many machine learning researches take the assumption that training and test distributions are identical (Figure~\ref{fig:generalization_types} left).
This means a main problem of conventional generalization is to learn a model working on the correct underlying distribution, and use it in the test.
In this case, the \textit{smoothness} assumption is important for learning the model, so that there exist some types of general purpose regularization, such as $L_2$ regularization and dropout.
Also, when we have more training data, we are likely to have better test performance.

Out-of-distribution (o.o.d.) generalization~\cite{bengio2017consciousness}, however, has different training and test distributions (Figure~\ref{fig:generalization_types} middle).
We focus on the test distribution manifold not in the training manifold, because the overlapping part is similar to conventional generalization.
The difference of the distributions needs both training and test distributions to define, so that training distribution alone does not have information for the difference.
This means the distribution difference information can only be given as prior knowledge during training, which is not general, but is specific for the test distribution.
So, in this case, more training data or general regularization does not directly help learning the distribution difference.
%
We are familiar with arguments in conventional generalization, but some of them may not directly apply in o.o.d. generalization.


\textit{Compositional generalization}, a.k.a. systematic generalization, is a type of o.o.d. generalization.
It has multiple components, and the generalization requires recombining values of different components in a novel way.
The values in each component appear in the training.
In Figure~\ref{fig:generalization_types} (right), a test sample is not in the training distribution, but when we decompose it to horizontal and vertical directions, the values of each component are in the training distribution, and we can combine these values for the test sample.
However, the components might be mixed together, and it is not straightforward to separate them.
This means we do not know the horizontal and the vertical directions in such cases.
When the representation has these orthogonal directions, we say it is a disentangled representation.


\subsection{Disentangled representations}
\begin{figure}[!ht]
\centering
\includegraphics[width=0.8\textwidth]{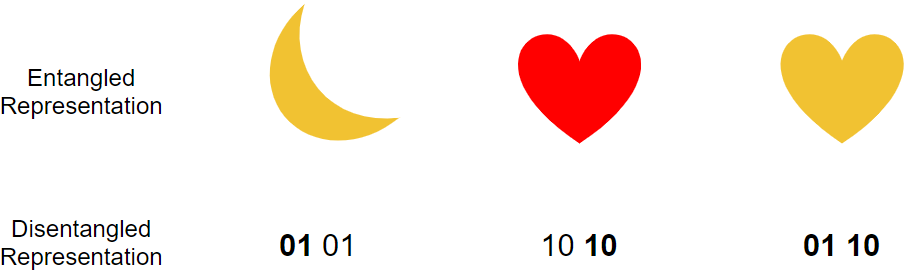}
\caption{Entangled (upper) and disentangled (lower) representations.}
\label{fig:disentangled_representations}
\end{figure}

A \textit{disentangled representation}~\cite{bengio2013deep} contains several \textit{separate} component representations.
Each component representation corresponds to an underlying component, or generative factor.
When a representation is not disentangled, it is an \textit{entangled representation}.
%
In the examples in Figure~\ref{fig:disentangled_representations}, we suppose to know that the components are color and shape.
The upper images are entangled representations, where color and shape are in the same image.
The lower vectors are disentangled representations, where the left vector is for color and the right vector is for shape.

Then we have several questions.
Where the types of components are from?
In this case, it means how do we know the components are color and shape?
Another question is what’s the relation between two representations, such as between the entangled and the disentangled representations?


\subsection{Subjectivity of components}
\label{sec:subjectivity_of_components}
We discuss the first question of where the types of components are from.
This part is still controversial and maybe not straightforward to agree, but we like to share the idea.
The idea is that the components can be subjectively defined by humans.
Sometimes they are common agreement of humans.
This also enables discussing different components in the same machine learning framework. We study a general way to encode human’s understanding of components to models.

Components can be subjective, but not arbitrary.
They are defined according to how humans perceived the world.
This means some components may be factors in real world physics, such as position and rotation.
They influence human perceptions, but humans decide the components.

In the example in Figure~\ref{fig:example_myth}, we have Sphinx and Centaur.
Though they are both created by compositional generalization, they have different components, one for face, and the other for upper body.
%
%
%
Another example is color.
We often use primary colors red, green and blue as components for colors.
However, their essential difference is the light wavelength.
There are three colors because generally humans have three types of photopsin proteins in eyes, each absorbing a primary color~\cite{solomon2007machinery}.
This means if an animal or a machine has four types of proteins, then they may have four primary colors.
So primary colors are not completely objective, and not completely subjective.
It depends on the objective biological mechanism of humans.


These are the examples of subjectivity of components.
Since machines are not humans, they do not know what subjective components humans have.
This means we need a general way to encode human’s understanding into models as prior knowledge.
%

\subsection{Conditional independence property}
\label{sec:conditinoal_independence_property}
\begin{figure}[!ht]
\centering
\subfloat[
Left only.
]{
\includegraphics[width=0.25\textwidth]{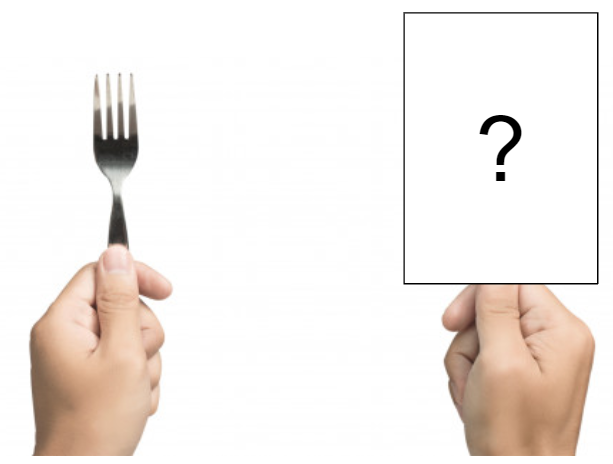}
\label{fig:cip_questions1}
}
\quad\quad\quad
\subfloat[
Both.
]{
\includegraphics[width=0.25\textwidth]{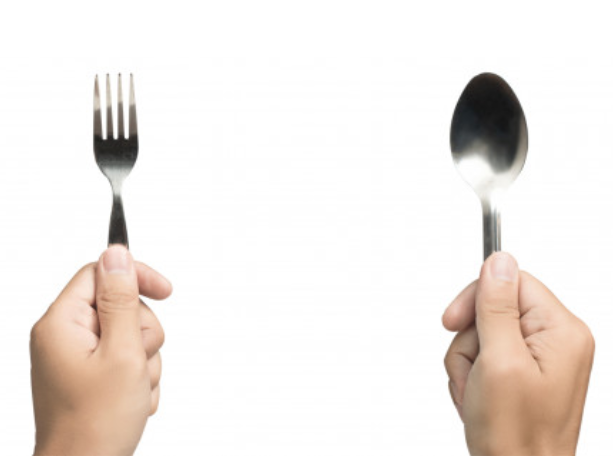}
\label{fig:cip_questions2}
}
\quad\quad\quad
\subfloat[
Right only.
]{
\includegraphics[width=0.25\textwidth]{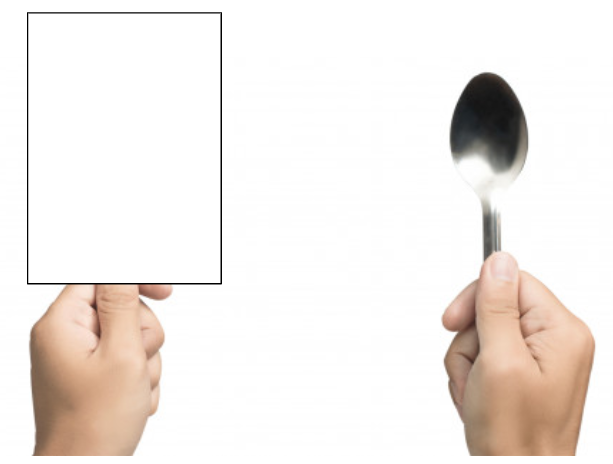}
\label{fig:cip_questions3}
}
\caption{
Conditional independence property.
What is in the right hand?
}
\label{fig:cip_questions}
\end{figure}

Let’s look at the relation between representations.
The key idea is conditional independence property.
We may first look at a question in Figure~\ref{fig:cip_questions1}.
What is in the right hand, when we see there is a fork in the left hand?
We do not know the exact answer, but the fork tells something about the answer.
We may guess the right hand has a knife or spoon.

Let’s ask again the question when we also have observation of the right hand (Figure~\ref{fig:cip_questions2}).
Given the observation of a spoon, we can tell the right hand has a spoon.
Then, let’s hide the left hand (Figure~\ref{fig:cip_questions3}).
In this case, the answer is the same, and hiding the left hand does not influence the answer.
This means the answer depends only on the observation of the right hand, though the left hand is related.
In other words, given the observation of the right hand, the answer is conditionally independent of other things.
This property is called \textit{conditional independence property}~\cite{li2021gradient}.


We formalize this property and find how it helps compositional generalization.
We consider two representations $X=X_1, \dots, X_K$ and $Y=Y_1,\dots,Y_K$.
They both have $K$ components, and each pair of components are aligned.
Conditional independence property can be summarized as $Y_i$ depends only on $X_i$.
This can be written in probability.
\begin{align*}
    & \forall i: P(Y_i | X_1, \dots, X_K, Y_1, \dots, Y_{i-1}, Y_{i+1}, \dots, Y_K) = P(Y_i | X_i).
\end{align*}

We also formalize compositional generalization (Figure~\ref{fig:generalization_types} right).
We consider a particular test sample with values of $X$ and $Y$.
In the training, each component value of $X_i$ appears, but the value of $X$ does not appear.
Note that this means the components are not marginally independent.
When the value of $X_i$ appears, the value of $Y_i$ has a high probability.
In the test, the value of $X$ appears, and we hope the predicted conditional probability of $Y$ given $X$ is high.
For example in Figure~\ref{fig:disentangled_representations}, a test sample can be a yellow heart, and it does not appear in training.
However, yellow appears in yellow moon, and heart appears in red heart in training.
$X$ is image, and $Y$ is label pair.
\begin{align*}
    & \text{In train,} && \text{In test,} \\
    & \forall i: P(X_i) > 0 , P(X_1, \dots, X_K) = 0, && P(X_1, \dots, X_K) > 0, \\
    & \forall i:P(Y_i | X_i) \text{ is high.} && P(Y_1, \dots, Y_K | X_1, \dots, X_K) \text{ is predicted high.}
\end{align*}
The conditional independence property bridges training and test distributions.
We first apply chain rule, and use compositional independence property.
\begin{align*}
    & P(Y_1, \dots, Y_K | X_1, \dots, X_K) = \prod^K_{i=1}P(Y_i | X_1, \dots, X_K, Y_1, \dots, Y_{i-1}) = \prod^K_{i=1}P(Y_i | X_i).
\end{align*}
When $P(Y_i | X_i)$ are all high, their product is high, so that $P(Y_1, \dots, Y_K | X_1, \dots, X_K)$ is high.
Therefore, a model satisfying conditional independence property addresses compositional generalization.


\section{Architecture design and training}
In this section, we introduce our setting, and describe an approach for compositional generalization.
We mainly discuss how to encode the prior knowledge and enable conditional independence property.

\subsection{Settings}
\label{sec:settings}

\begin{wrapfigure}{r}{0.35\linewidth}
  \centering
    \begin{tikzpicture}
    \path [every node/.style={minimum width=0cm, minimum height=1.3cm]}]
      node (a) at (0,0) {$X$}
      [xshift=2.0cm]
      node (b) at (0,0) {$H$}
      [xshift=2.0cm]
      node (c) at (0,0) {$Y$}
      ;
      \draw[-latex] (a.east) -- node [above] {\small Encode} (b.west) ;
      \draw[-latex] (b.east) -- node [above] {\small Decode} (c.west) ;
    \end{tikzpicture}
  \caption{Problem setting. Both input $X$ and output $Y$ are entangled. We use a disentangled hidden representation $H$. The model has encoding and decoding modules.}
  \label{fig:problem_setting}
\end{wrapfigure}
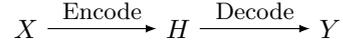

We focus on a general setting for compositional generalization.
We consider a problem with both entangled input $X$ and entangled output $Y$, and components are aligned.
For example in language translation, both input and output languages are entangled with grammar and lexicon.
The input grammar decides output grammar, and input lexicon decides output lexicon.

Compositional generalization requires recombining values of different components in a novel way.
As we consider component types are subjective (Section~\ref{sec:subjectivity_of_components}), it requires knowing what are the types of components, such as shape and color.
%
There are different ways to add this prior knowledge.
In some cases, the prior knowledge is in the design of data structure (position in image). Some approaches design training data distribution to make the components statistically marginally independent.
We attend to using the prior knowledge in model architecture design with particular regularization.


We focus on using disentangled representation, because it is conceptually straightforward for compositional generalization.
We do not assume statistical independence between components or use annotations on components.
In such a setting, we have encoding and decoding modules (Figure~\ref{fig:problem_setting}).
Encoder converts entangled input $X$ to hidden disentangled representation $H$, and decoder converts $H$ to entangled output $Y$.
We can set $Y=X$ for unsupervised representation learning.

\subsection{Strategy}
We hope to enable conditional independence property (Section~\ref{sec:conditinoal_independence_property}) by encoding prior knowledge for components.
This means we expect a component representation $H_i$ to have exact information of the corresponding component.
For example, we hope a component representation contains the color information.
This requires controlling information of a random variable (component representation).

%
To achieve it, we hope to design a loss function that has the minimum value when the information is expected.
So we study the relation between optimization loss and entropy of a component representation.
Note that entropy measures the amount of information, not the contents, but we use entropy for intuitive explanation.
%
%
%
Also note that we consider a (multi-dimensional) representation as a random variable.
We discuss the distribution, and its entropy, of this random variable with all the samples in a dataset.
This means for one dataset, we have only one distribution for the component representation and only one entropy for the distribution.
%

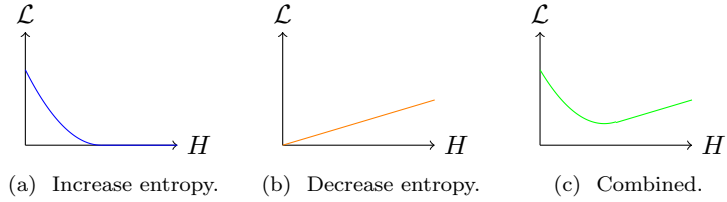
\begin{figure}[!ht]
\centering
\subfloat[
Increase entropy.
]{
\begin{tikzpicture}
  \draw[->] (0, 0) -- (2, 0) node[right] {$H$};
  \draw[->] (0, 0) -- (0, 1.5) node[above] {$\mathcal{L}$};
  \draw[domain=0:1, smooth, variable=\x, blue] plot ({\x}, {(\x-1)*(\x-1)});
  \draw[domain=1:2, smooth, variable=\x, blue] plot ({\x}, {0});
\end{tikzpicture}
\label{fig:electromagnetic_energy}
}
\quad
\subfloat[
Decrease entropy. 
]{
\begin{tikzpicture}
  \draw[->] (0, 0) -- (2, 0) node[right] {$H$};
  \draw[->] (0, 0) -- (0, 1.5) node[above] {$\mathcal{L}$};
  \draw[domain=0:2, smooth, variable=\x, orange] plot ({\x}, {0.3*\x});
\end{tikzpicture}
\label{fig:gravity}
}
\quad
\subfloat[
Combined.
]{
\begin{tikzpicture}
  \draw[->] (0, 0) -- (2, 0) node[right] {$H$};
  \draw[->] (0, 0) -- (0, 1.5) node[above] {$\mathcal{L}$};
  \draw[domain=0:1, smooth, variable=\x, green] plot ({\x}, {(\x-1)*(\x-1)+0.31*\x});
  \draw[domain=1:2, smooth, variable=\x, green] plot ({\x}, {0.3*\x});
\end{tikzpicture}
\label{fig:superposition}
}
\caption{
Loss and entropy.
Horizontal axis is component entropy $H(H_i)$.
Vertical axis is loss.
}
\label{fig:potential_energy}
\end{figure}

The strategy is to design a convex loss with the minimum at the target entropy (Figure~\ref{fig:superposition}, and note that the horizontal axis is entropy instead of parameters).
This requires techniques to increase entropy, decrease entropy, and enable local turning of loss at the target entropy (locality).
We look at related techniques in machine learning (Table~\ref{table:machine_learning_techniques}).
Prediction loss increases entropy, because when we train a model to have correct prediction, an intermediate representation should contain more information to do that.
During increasing entropy, we can encode local turning point by architecture design as we will discuss in Section~\ref{sec:architecture_design}.
Regularization can reduce entropy, and we discuss in Section~\ref{sec:entropy_regularization}.
However, it is not clear how to encode locality during reducing entropy.

\begin{table}[!ht]
\caption{Machine learning techniques.}
\label{table:machine_learning_techniques}
\begin{center}
\begin{tabular}{lll}
 & Increase entropy & Decrease entropy \\
\hline
Loss & \textcolor{blue}{Prediction loss} & \textcolor{orange}{Regularization} \\
Locality & \textcolor{blue}{Architecture design} & Not clear
\end{tabular}
\end{center}
\end{table}

With the above availability, we design two losses.
For the loss to increase entropy (Figure~\ref{fig:electromagnetic_energy}), we use prediction loss and architecture design.
The loss rapidly decreases when entropy increases and is below the target value, and it is constant after that.
For the loss to decrease entropy (Figure~\ref{fig:gravity}), we use regularization.
The loss stably increases as entropy increases.
They together form the expected curve (Figure~\ref{fig:superposition}).
This approach has two advantages.
First, the target position is encoded only to one loss.
Second, it does not need specific values for the losses.

\subsection{Architecture design and prediction loss}
\label{sec:architecture_design}

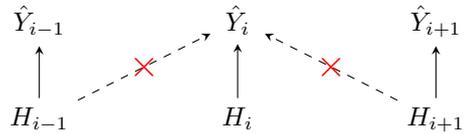
\begin{wrapfigure}{r}{0.45\linewidth}
\centering
\begin{tikzpicture}
  \matrix (m) [matrix of math nodes,row sep=2em,column sep=5em,minimum width=2em]
  {
     \hat{Y}_{i-1} & \hat{Y}_i & \hat{Y}_{i+1} \\
     H_{i-1} & H_i & H_{i+1} \\};
  \path[-stealth]
    (m-2-1) edge (m-1-1)
    (m-2-1) edge[dashed] node {\textcolor{red}{\Large$\times$}} (m-1-2)
    (m-2-2) edge (m-1-2)
    (m-2-3) edge[dashed] node {\textcolor{red}{\Large$\times$}} (m-1-2)
    (m-2-3) edge (m-1-3);
\end{tikzpicture}
\caption{
  Architecture design.
  Avoiding connections from $H_j (j\neq i)$ to $\hat{Y}_i$, so only $H_i$ can influence $\hat{Y}_i$.
  The prediction loss makes $H_i$ at least contain information of $Y_i$.
}
\label{fig:structure_design}
\end{wrapfigure}

We first discuss how to encode locality when increasing information.
In Figure~\ref{fig:electromagnetic_energy}, the loss decreases when entropy is not enough, and the loss is constant when entropy is enough.
This means that we hope to make a component representation have at least certain information.
We achieve this by architecture design combined with prediction loss.

When each output component $\hat{Y}_i$ is connected only to one corresponding hidden component representation $H_i$, information of $\hat{Y}_i$ can come only from this hidden component representation (Figure~\ref{fig:structure_design}).
Note that this also means $H_i$ is connected forward only to $\hat{Y_i}$.
We consider that if the output $\hat{Y}$ is correct, all its components $\hat{Y}_i$ should be correct.
Since this component needs to be correct when reducing prediction loss, the hidden component representation should contain at least the information of the component.
Please also refer to Appendix~\ref{sec:extended_topic} for extended discussions.


\begin{wrapfigure}{l}{0.5\linewidth}
\centering
\subfloat[
Hidden components.
]{
\includegraphics[height=0.18\textwidth]{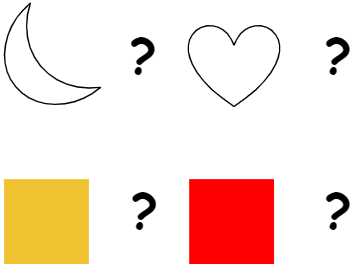}
}
\subfloat[
Entangled outputs.
]{
\includegraphics[height=0.18\textwidth]{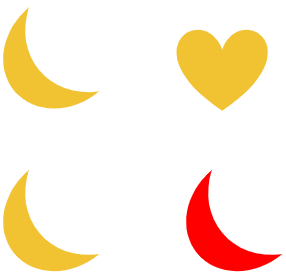}
}
\caption{
Example of architecture design effect.
The upper row is for the shape component, and the lower row is for the color component.
One output component changes its value only when the corresponding hidden component representation changes its value.
To make the output correct, each hidden representation should at least contain the information for the component.
}
\label{fig:architecture_design_example}
\end{wrapfigure}

This is the way that we encode component prior knowledge in the architecture design, i.e., we describe how it generates output.
Note that this generating process might be different from the real generating process.
For example, we have generative factors of shape, size and color for an apple.
For a machine, a generating process can first choose a shape, then adjust the size and paint color.
However, a real apple grows with these three components changing together.

This technique works for the decoding process, because the output needs to be compared with ground truth to compute the loss.
Also, for humans, describing the decoding process is easier than the encoding process.
For example, computer graphics is easier than computer vision.
Computer graphics, in many cases, does not need machine learning, such as developing 3D games.
However, computer vision is hard without machine learning.

Let’s look at an example in Figure~\ref{fig:architecture_design_example}.
There are two component representations, one for shape and the other for color.
We can design an architecture to achieve the following effects.
The output shape changes only when the first component representation changes its value.
The output color changes only when the second component representation changes its value.
With such design, to produce correct output, the first component representation should at least contain the shape information, and the second component representation should at least contain the color information, because the other representation is not able to provide the information.
Please refer to \cite{li2021necessary} for more analysis.

\subsection{Entropy regularization}
\label{sec:entropy_regularization}

\begin{wrapfigure}{r}{0.35\linewidth}
\centering
\includegraphics[width=0.22\textwidth]{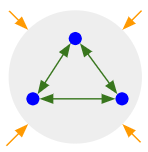}
\caption{
Entropy Regularization.
\textcolor{darkgreen}{Green} is for noise.
\textcolor{orange}{Orange} is for norm regularization.
}
\label{fig:entropy_regularization}
\end{wrapfigure}

We then talk about reducing the information of a component representation.
This does not require component specific prior knowledge.
Entropy for a random variable can be roughly understood as the number of possible values.

Entropy regularization~\cite{li2019compositional} aims at reducing entropy of a component representation.
Given a representation $x$, we compute the $L_2$ norm and add normal noise to each element of the representation.
This decreases the channel capacity, so that the entropy for the representation reduces.
We then feed the noised representation to the next layer, and add the norm to loss function.
\begin{align*}
    & \mathcal{L} = \mathcal{L}_\text{original} + \lambda L_2(x)
    && \text{EntReg}(x) = x + \alpha \mathcal{N}(0, I)
\end{align*}
where $\alpha$ is a weight of noise, positive for training and zero for inference.
$\lambda$ is a coefficient.

Please see Figure~\ref{fig:entropy_regularization} for intuitive illustration.
The noise makes different values far from each other in vector space.
If they are close, the noise will make them not distinguishable, so the prediction would be wrong.
At the same time, norm regularization makes different values close to each other to reduce the region of manifold.
These two forces squash the values, so that unnecessary values will be merged.
With less number of possible values, the entropy reduces.
Please also refer to Appendix~\ref{sec:entropy_language}.

\subsection{Stochastic sampling and gradient descent}
We discussed two losses to increase and decrease entropy.
However, during the optimization of neural networks, there are other influences acting like losses, and we consider them as losses for simple explanation.
These losses come from stochastic gradient descent.
It is a widely used optimization algorithm with many variations, and the following arguments apply to them.

One loss is from stochastic sampling.
This reduces entropy because it adds noise.
The effect is similar to entropy regularization, but this is weak.
This effect appears mainly during the later stage of training.
Occasionally, this enables learning compositionality without entropy regularization.
For more details, please refer to \cite{shwartz2017opening}.

Another loss comes from gradient descent.
It increases entropy of a component representation.
This is because the optimization process imposes a bias toward non-compositional solutions, which is because gradient seeks the steepest direction, so that it uses all available and redundant input information.
This mainly happens during the early stage of training.
This effect can be canceled by entropy regularization, so entropy regularization is important.
%
Note that it is not prominent when there is only one solution, e.g., with linear model.
%
Please refer to \cite{li2021gradient} for more details.



\subsection{Summary of four losses}
\begin{figure}[!ht]
\centering
\subfloat[
Prediction loss (\textcolor{blue}{blue}).
]{
\begin{tikzpicture}
  \draw[->] (0, 0) -- (2, 0) node[right] {$H$};
  \draw[->] (0, 0) -- (0, 1.5) node[above] {$\mathcal{L}$};
  \draw[domain=0:1, smooth, variable=\x, blue] plot ({\x}, {(\x-1)*(\x-1)});
  \draw[domain=1:2, smooth, variable=\x, blue] plot ({\x}, {0});
\end{tikzpicture}
\begin{tikzpicture}
  \draw[->] (0, 0) -- (2, 0) node[right] {$H$};
  \draw[->] (0, 0) -- (0, 1.5) node[above] {$\mathcal{L}$};
  \draw[domain=0:1, smooth, variable=\x, green] plot ({\x}, {(\x-1)*(\x-1)});
  \draw[domain=1:2, smooth, variable=\x, green] plot ({\x}, {0});
\end{tikzpicture}
}
\quad\quad
\subfloat[
Stochastic sampling (\textcolor{red}{red}).
]{
\begin{tikzpicture}
  \draw[->] (0, 0) -- (2, 0) node[right] {$H$};
  \draw[->] (0, 0) -- (0, 1.5) node[above] {$\mathcal{L}$};
  \draw[domain=0:1, smooth, variable=\x, blue] plot ({\x}, {(\x-1)*(\x-1)});
  \draw[domain=1:2, smooth, variable=\x, blue] plot ({\x}, {0});
  \draw[domain=0:2, smooth, variable=\x, red] plot ({\x}, {0.1*\x});
\end{tikzpicture}
\begin{tikzpicture}
  \draw[->] (0, 0) -- (2, 0) node[right] {$H$};
  \draw[->] (0, 0) -- (0, 1.5) node[above] {$\mathcal{L}$};
  \draw[domain=0:1, smooth, variable=\x, green] plot ({\x}, {(\x-1)*(\x-1)+0.1*\x});
  \draw[domain=1:2, smooth, variable=\x, green] plot ({\x}, {0.1*\x});
\end{tikzpicture}
}
\\
\subfloat[
Gradient descent (\textcolor{cyan}{cyan}).
]{
\begin{tikzpicture}
  \draw[->] (0, 0) -- (2, 0) node[right] {$H$};
  \draw[->] (0, 0) -- (0, 1.5) node[above] {$\mathcal{L}$};
  \draw[domain=0:1, smooth, variable=\x, blue] plot ({\x}, {(\x-1)*(\x-1)});
  \draw[domain=1:2, smooth, variable=\x, blue] plot ({\x}, {0});
  \draw[domain=0:2, smooth, variable=\x, red] plot ({\x}, {0.1*\x});
  \draw[domain=0:2, smooth, variable=\x, cyan] plot ({\x}, {0.2*(2.5-\x)});
\end{tikzpicture}
\begin{tikzpicture}
  \draw[->] (0, 0) -- (2, 0) node[right] {$H$};
  \draw[->] (0, 0) -- (0, 1.5) node[above] {$\mathcal{L}$};
  \draw[domain=0:1, smooth, variable=\x, green] plot ({\x}, {(\x-1)*(\x-1)+0.1*\x+0.2*(2.5-\x)});
  \draw[domain=1:2, smooth, variable=\x, green] plot ({\x}, {0.1*\x+0.2*(2.5-\x)});
\end{tikzpicture}
}
\quad\quad
\subfloat[
Entropy regularization (\textcolor{orange}{orange}).
]{
\begin{tikzpicture}
  \draw[->] (0, 0) -- (2, 0) node[right] {$H$};
  \draw[->] (0, 0) -- (0, 1.5) node[above] {$\mathcal{L}$};
  \draw[domain=0:1, smooth, variable=\x, blue] plot ({\x}, {(\x-1)*(\x-1)});
  \draw[domain=1:2, smooth, variable=\x, blue] plot ({\x}, {0});
  \draw[domain=0:2, smooth, variable=\x, red] plot ({\x}, {0.1*\x});
  \draw[domain=0:2, smooth, variable=\x, cyan] plot ({\x}, {0.2*(2.5-\x)});
  \draw[domain=0:2, smooth, variable=\x, orange] plot ({\x}, {0.3*\x});
\end{tikzpicture}
\begin{tikzpicture}
  \draw[->] (0, 0) -- (2, 0) node[right] {$H$};
  \draw[->] (0, 0) -- (0, 1.5) node[above] {$\mathcal{L}$};
  \draw[domain=0:1, smooth, variable=\x, green] plot ({\x}, {(\x-1)*(\x-1)+0.1*\x+0.2*(2.5-\x)+0.3*\x});
  \draw[domain=1:2, smooth, variable=\x, green] plot ({\x}, {0.1*\x+0.2*(2.5-\x)+0.3*\x});
\end{tikzpicture}
}
\caption{
Four losses.
Horizontal axis is entropy $H(H_i)$ for a component representation on training distribution.
Vertical axis is loss $\mathcal{L}$.
In each pair of figures, left is individual loss, and right is summed loss (\textcolor{green}{green}).
The summed loss for all influences has the minimum close to the expected point.
}
\label{fig:forces}
\end{figure}
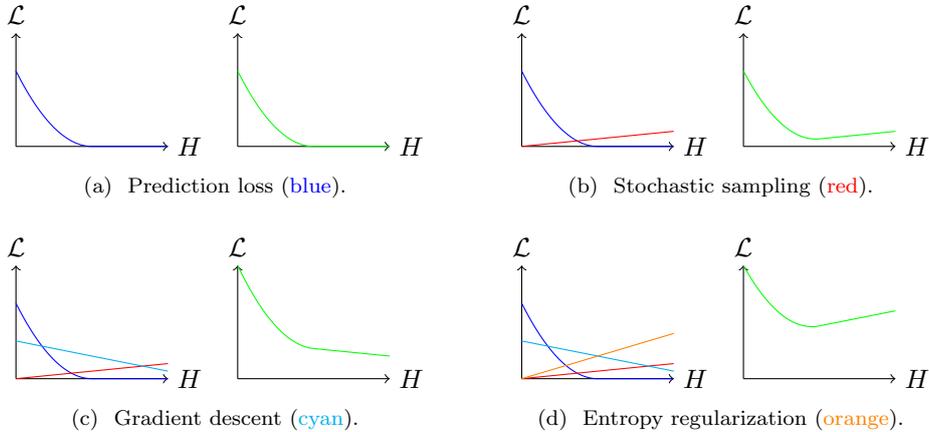

Let’s summarize the four losses during optimization.
Please see Figure~\ref{fig:forces} for intuitions.
%
The first loss is from prediction loss with architecture design.
This loss decreases rapidly when entropy is small, and it is constant after the entropy is above the target value.
The second loss is from stochastic sampling~\cite{shwartz2017opening}.
This exists naturally but is weak.
The third loss is from gradient descent~\cite{li2021gradient}.
The fourth loss is from entropy regularization~\cite{li2019compositional}, and it counteracts the effect from gradient descent.
This loss should be less steep than the prediction loss, so that the summed loss has the lowest point near to the expected value.

\section{Inference}
In this section, we look at problems during inference.
We also discuss conjectures for human behaviors (Appendix~\ref{sec:conjectures}).
We then analyze that language tasks are less likely to suffer from the problems in inference (Appendix~\ref{sec:inference_language}).

\subsection{Problem}
So far, we have discussed learning compositionality during training.
However, our goal is compositional generalization, and we hope for high performance in the test.
So a question is whether the model still works on test distribution.
We have both encoding and decoding parts (Figure~\ref{fig:problem_setting}).

Decoding still works if encoding is correct.
This is because of the architecture design, where only the corresponding component representation produces the component in output.
Since the component representation is correct, and it is in the same manifold as training by definition of compositional generalization, the network produces a correct output component.

However, the encoding part may not work on test distribution.
It extracts disentangled representation from entangled input representation.
This extraction network, however, can be a general network, and we do not have special treatment for it.
By definition of compositional generalization, the input manifold changes, and a general network does not work well in such cases.
Therefore, the encoding network may not produce correct disentangled representation.
Please also refer to \cite{li2021transferability}.


\subsection{Solution}
\begin{figure}[!ht]
  \centering
  \subfloat[Training flowchart. The three modules are trained with end-to-end optimization.]{
    \begin{tikzpicture}
    \path [every node/.style={minimum width=0cm, minimum height=1.3cm]}]
      node (a) at (0,0) {$X$}
      [xshift=2.0cm]
      node (b) at (0,0) {$H$}
      [xshift=2.0cm]
      node (c) at (0,0) {$Y$}
      ;
      \draw[-latex] ([yshift=0.07 cm]a.east) -- node [above] {\small$g(X;\phi)$} ([yshift=0.07 cm]b.west) ;
      \draw[latex-] ([yshift=-0.07 cm]a.east) -- node [below] {\small$h(H;\psi)$} ([yshift=-0.07 cm]b.west) ;
      \draw[-latex] (b.east) -- node [above] {\small$f(H;\theta)$} (c.west) ;
    \end{tikzpicture}
  }
  \hspace{2mm}
  \subfloat[Inference flowchart.
  (Left) initial hidden representations extraction.
  (Middle) optimization of hidden representations as module input.
  (Right) output prediction.]{
    \begin{tikzpicture}
    \path [every node/.style={minimum width=0cm, minimum height=1.3cm]}]
      node (a) at (0,0) {$X$}
      [xshift=2.0cm]
      node (b) at (0,0) {$H$}
      [xshift=0.75cm]
      node (c) at (0,0) {$X$}
      [xshift=2.0cm]
      node (d) at (0,0) {$H$}
      [xshift=0.75cm]
      node (e) at (0,0) {$H$}
      [xshift=2.0cm]
      node (f) at (0,0) {$Y$}
      ;
      \draw[-latex] (a.east) -- node [above] {\small$g(X;\phi)$} (b.west) ;
      \draw[latex-] (c.east) -- node [below] {\small$h(H;\psi)$} (d.west) ;
      \draw[-latex] (e.east) -- node [above] {\small$f(H;\theta)$} (f.west) ;
    \end{tikzpicture}
  }
  \caption{
  Flowcharts of the proposed approach.
  $X$ is input, $Y$ is output, and $H$ is hidden representation.
  The architecture has three modules: $g,h,f$.
  }
  \label{fig:architecture}
\end{figure}
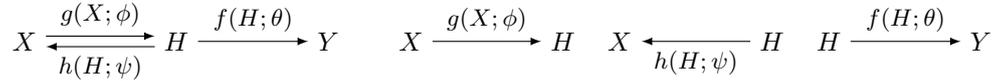

One idea to address this problem is to convert the encoding problem to a decoding problem by reversing input and output and specify architecture design with an additional decoding network $h$ (Figure~\ref{fig:architecture}).
Similar to the other decoding network $f$, $h$ works when each component is in its training manifold.
Since the input and the output is opposite, we cannot get the hidden representation $H$ with a forward pass.
So we use optimization to get the input $H$ that best produces the output $X$.

To regularize each test $H_i$ in its training manifold, 
we may keep the manifold information, and use it in test.
A straightforward way to keep the information is to store some training samples.
They may be stored as input representation or hidden representation.
In test, we make each test $H_i$ close to the corresponding training ones.
The encoding network $g$ provides initial hidden representations.

In summary (Figure~\ref{fig:architecture}), we jointly train three models $g,h,f$ in training.
In inference, we first use the encoder $g$ to get initial hidden representation.
Then we use the additional decoder $h$ to optimize the hidden representation to reconstruct the input with manifold regularization.
We then use the original decoder $f$ to convert the optimized hidden representation to output.

%
%
%
%

\section{Conclusion}
This report introduces compositional generalization and an approach to it, with pointers to a series of corresponding papers.
%
%
%
It has three important key points.
First, what is compositional generalization.
It is an out-of-distribution generalization with recombination of seen component values in a novel way.
Second, conditional independence property.
This is the core property of compositional generalization.
It means an output component depends only on the corresponding input component.
The last point is controlling random variable information.
It enables conditional independence property.
We achieve it by squeezing entropy from above and below.
We hope this report will help understanding compositional generalization and advancing artificial intelligence.

\section*{Acknowledgments}
We thank Mohamed Elhoseiny, Liang Zhao, Wei Xu, Kenneth Church, Joel Hestness, Jianyu Wang, Yi Yang and Zhuoyuan Chen for helpful suggestions and discussions.

\bibliographystyle{unsrt}
\bibliography{main}
\appendix

\section{Partial observation of output combinations}
\label{sec:extended_topic}
In Section~\ref{sec:architecture_design}, we discussed architecture design that enables a component representation to contain at least the information of a component.
One condition here is that when prediction $\hat{Y}$ equals to ground-truth $Y$, all the component outputs are correct $\hat{Y}_i = Y_i$.
%
As an extended topic, in some complicated cases, the condition may not be met when the same output $Y$ can ambiguously correspond to different combinations of component values, if it discards a part of information for the combinations.

However, even in such cases, the arguments still hold with disambiguation.
Broadly speaking, how to disambiguate is another type of prior knowledge.
For example, reducing entropy of each component representation makes the ambiguity disappear in some tasks.
Then, the entropy regularization also performs disambiguation.
This mechanism is used in \cite{li2019compositional}, where a combination is for syntax tree and words on nodes, but $Y$ only contains words.


\section{Entropy regularization in language learning}
\label{sec:entropy_language}
We like to share a joke for ``law of entropy increase'' in human language learning.
Law of entropy increase originally says in an isolated system, the entropy increases over time.
%
Here, a beginner of learning a second language is likely to "overuse" compositional generalization to create unnatural phrases.
As one becomes more fluent over time, the problem is less (less highly compositional).
Since entropy reduction helps compositional generalization, this means entropy increases over time.

\section{Conjectures for system 1 and system 2 cognition}
\label{sec:conjectures}
\begin{figure}[!ht]
\centering
\subfloat[
System 1: fast
]{
\includegraphics[height=0.22\textwidth]{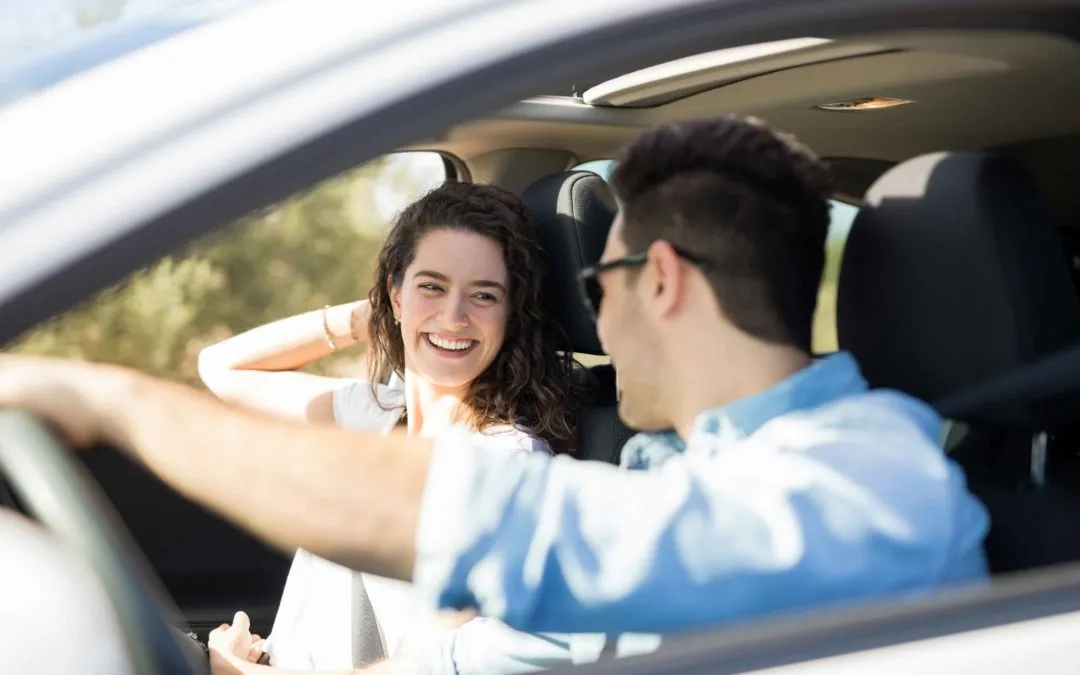}
}
\quad\quad
\subfloat[
System 2: slow
]{
\includegraphics[height=0.22\textwidth]{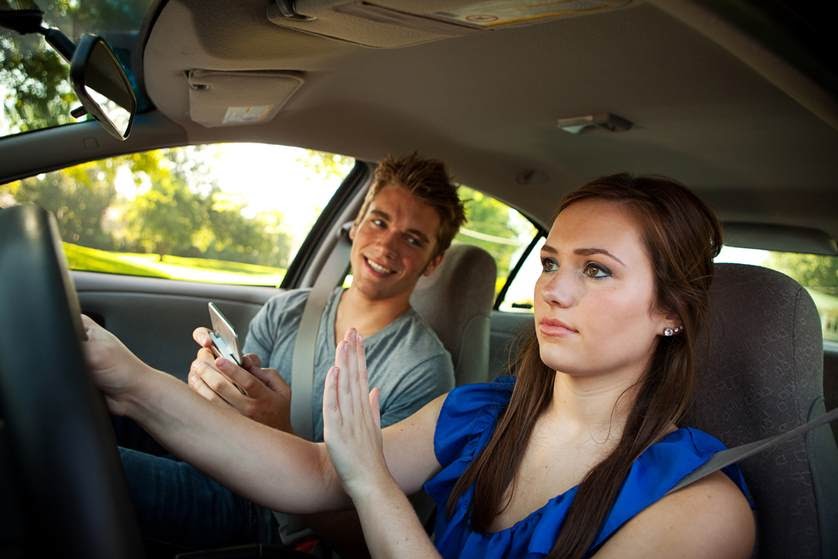}
}
\caption{Examples for human cognition.
}
\label{fig:system_examples}
\end{figure}

We like to share some conjectures for system 1 and system 2 cognition~\cite{kahneman2011thinking}.
System 1 is a fast and unconscious cognition process.
System 2 is a slow and conscious cognition process.
Figure~\ref{fig:system_examples} is a borrowed example (\url{https://youtu.be/4KpZBiKda0k}).
System 1 is driving on a familiar road.
The driver is relaxed, and can drive while chatting.
System 2 is driving on an unfamiliar road.
The driver needs to focus on driving.

In system 2, humans need more time and attention.
What do we do with these resources?
The conjecture is we are doing optimization.
More precisely, when the input is familiar, the encoding network works well, so that optimization is simple and fast (maybe fewer optimization steps).
When the input is unfamiliar, the encoding network does not provide a good initial hidden representation, so the optimization is difficult and slow.

Another conjecture is that our long-term memory is used for manifold regularization, which requires storing training samples.

\section{Inference for language}
\label{sec:inference_language}

\begin{wrapfigure}{r}{0.5\linewidth}
\centering
\includegraphics[width=0.45\textwidth]{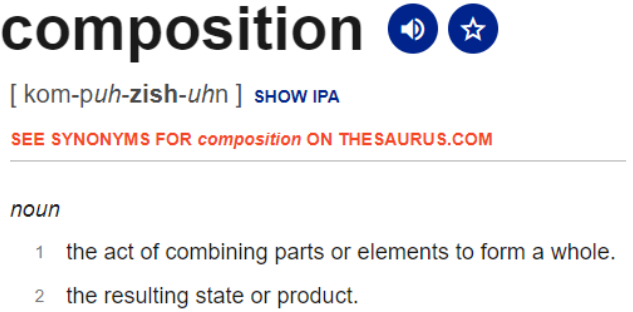}
\caption{
A word ``composition'' in dictionary (\url{www.dictionary.com}).
}
\label{fig:dictionary}
\end{wrapfigure}

When we read new articles, such as news, there are many new sentences.
If sentence structures are simple, we can often read at a constant speed.

This phenomena might be explained by transferable units, such as words.
In different situations, a word is likely to have the same word extraction information, syntactic information and semantic information.
These types of information are so stable that we can create a general purpose dictionary (e.g.,  Figure~\ref{fig:dictionary}).
The extraction information is the spell, the syntactic information is part-of-speech and the semantic information is the explanations.
Note that the information might not be deterministic, but the entries are stable.
This means we can find words in the same way in different situations, and use the same values for each component by concatenating word level component representations.
Because of these properties, we avoid complicated optimization during inference, and have constant speed in some language tasks.

%

\end{document}